\documentclass[cha, amsmath, amssymb, reprint]{revtex4-1}

\usepackage{graphicx}
\usepackage{physics, amsmath, amssymb}
\usepackage{booktabs}
\usepackage{multirow}
\usepackage[bookmarks=true, hidelinks, hyperindex, breaklinks]{hyperref}

\begin{document}

\title[Working title]{Hybridizing Traditional and Next-Generation Reservoir Computing \\ to Accurately and Efficiently Forecast Dynamical Systems}

\author{R. Chepuri}
    \affiliation{Department of Physics, University of Maryland, College Park, MD 20742}
    
\author{D. Amzalag}
    \affiliation{University of Chicago, Chicago, IL 60637}

\author{T.M. Antonsen}
    \affiliation{Department of Electrical and Computer Engineering, University of Maryland, College Park, MD 20742}
        \affiliation{Department of Physics, University of Maryland, College Park, MD 20742}
    \affiliation{Institute for Research in Electronics and Applied Physics (IREAP), College Park, MD 20742}

\author{M. Girvan}
    \email{girvan@umd.edu}
    \affiliation{Department of Physics, University of Maryland, College Park, MD 20742}
    \affiliation{Institute for Physical Science and Technology (IPST), College Park, MD 20742}
    \affiliation{Institute for Research in Electronics and Applied Physics (IREAP), College Park, MD 20742}
    \affiliation{Santa Fe Institute, Santa Fe, NM 87501}

\date{\today}

\begin{abstract}

Reservoir computers (RCs) are powerful machine learning architectures for time series prediction. Recently, next generation reservoir computers (NGRCs) have been introduced, offering distinct advantages over RCs, such as reduced computational expense and lower training data requirements. However, NGRCs have their own practical difficulties, including sensitivity to sampling time and type of nonlinearities in the data. Here, we introduce a hybrid RC-NGRC approach for time series forecasting of dynamical systems. We show that our hybrid approach can produce accurate short term predictions and capture the long term statistics of chaotic dynamical systems in situations where the RC and NGRC components alone are insufficient, e.g., due to constraints from limited computational resources, sub-optimal hyperparameters, sparsely-sampled training data, etc. Under these conditions, we show for multiple model chaotic systems that the hybrid RC-NGRC method with a small reservoir can achieve prediction performance approaching that of a traditional RC with a much larger reservoir, illustrating that the hybrid approach can offer significant gains in computational efficiency over traditional RCs while simultaneously addressing some of the limitations of NGRCs. Our results suggest that hybrid RC-NGRC approach may be particularly beneficial in cases when computational efficiency is a high priority and an NGRC alone is not adequate.
\end{abstract}


\maketitle

\textbf{
Predicting the behavior of a dynamical system over time poses a significant challenge, especially when dealing with chaotic or complex systems. Reservoir computing, a type of machine learning framework, has emerged as a promising solution for this task. It offers advantages over deep learning methods, particularly in terms of computational efficiency. However, harder prediction tasks generally require larger, more computationally expensive reservoir computers (RCs) containing numerous artificial neurons. To tackle this issue, researchers have introduced next-generation reservoir computers (NGRCs), which boast even greater computational efficiency. While NGRCs have shown remarkable performance across various scenarios, they sometimes struggle with tasks that traditional RCs handle easily. In this study, we propose a novel hybrid approach that leverages the strengths of both RCs and NGRCs. By combining a small, computationally efficient RC with an NGRC, our hybrid model is able to achieve the performance and flexibility of a much larger RC while still preserving a substantial portion of the efficiency advantages of an NGRC.
}

\section{Introduction}
\label{sec:intro}

\begin{figure*}
    \includegraphics{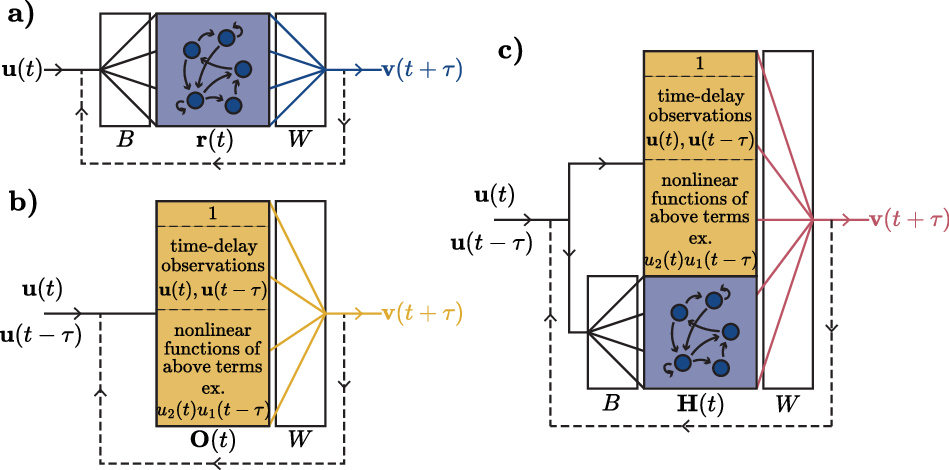}
    \caption{\label{fig:schematics} \textbf{a)} Reservoir computer (RC) schematic. Time series observations $\vb{u}(t)$ are fed into a high-dimensional reservoir with state $\vb{r}(t)$ via an input matrix $B$, then an output matrix $W$ is trained to predict the next data point in the series (i.e., at time $t + \tau$). Predictions at times $t > t_{\text{train}}$ are made by switching to autonomous mode in which outputs of the reservoir are repeatedly fed back in as input (dashed line). \textbf{b)} Next-generation reservoir computers (NGRCs) replace the reservoir with a nonlinear feature vector $\vb{O}(t)$ that is constructed using time-delayed observations. \textbf{c)} Our hybrid RC-NGRC prediction approach uses a hybrid feature vector $\vb{H}(t)$ that is the concatenation of a reservoir state with a NGRC feature vector in order to produce a prediction.}
\end{figure*}

Reservoir computing has emerged as a powerful machine learning architecture for forecasting dynamical systems \cite{jaeger_2001, maass.etal_2002, lukosevicius.jaeger_2009, pathak.etal_2018, lu.etal_2018, tanaka.etal_2019}. In a reservoir computer (RC), a high-dimensional nonlinear system called the reservoir is used to learn the flow of a dynamical system, and subsequently make a forecast. RCs have been shown not only to achieve impressive short term forecast accuracy, notably in the difficult case of chaotic systems, but also to reproduce the statistical properties (i.e., capture the ``climate'') of the true system in the long term \cite{lu.etal_2018}. Though RCs are relatively effective, several drawbacks have been noted, such as the need to tune many hyperparameters for optimal performance \cite{tanaka.etal_2019}.

An alternative to an RC, dubbed a next-generation reservoir computer (NGRC), has been introduced that avoids the use of a reservoir entirely \cite{gauthier.etal_2021}. Implemented as a kind of a nonlinear vector autoregression (NVAR) machine \cite{lutkepohl_2005}, an NGRC replaces the reservoir state vector with a feature vector that includes nonlinear functions of time-delayed observations of the dynamical system. NGRCs have been shown to be capable of forecasting several prototypical chaotic systems at greatly reduced computational cost compared to RCs, and are in fact mathematically equivalent to a variant of RCs with linear reservoir nodes and nonlinear readout \cite{bollt_2021}. However, implementations of NGRCs in practice show substantially different challenges from traditional RC implementations. For example, the presence of specific nonlinearities related to the true system in the NGRC feature vector has been shown to be essential for forecasting some dynamical systems \cite{zhang.cornelius_2023}. We also note that when data are sampled from a system at large time steps, nonlinearities associated with finite-time updates are no longer a good approximation of those of the sparsely sampled system, providing a challenge for NGRCs. Indeed, we find that NGRCs struggle to forecast prototypical chaotic systems when training data is sampled with a large time step, even though they offer very strong performance at small sampling time steps (Section \ref{sec:hybrid_improvement_large_timestep}).

In this paper, we introduce a hybrid RC-NGRC approach that leverages the strengths of both RCs and NGRCs. Previous work has shown that RCs can be hybridized with the output of a knowledge-based model \cite{pathak.etal_2018a} or the output of another machine-learning-based prediction scheme \cite{koster.etal_2023}. Here, rather than hybridize an RC with the \textit{output} of an NGRC, we utilize a hybrid feature/representation vector that is the concatenation of the reservoir state with \textit{all} the terms in a NGRC feature vector (Figure \ref{fig:schematics}c)), giving our hybrid more flexibility in allowing the RC component to counter limitations in the NGRC component. We find that for some forecasting tasks that particular RC and NGRC implementations struggle with, our hybrid RC-NGRC approach can substantially outperform them at both short term forecasting and long term `climate replication' (i.e. the ability of the predicted dynamics to match the statistical features of the true system even after short term forecasts are no longer accurate).

We demonstrate the utility of our approach with model chaotic systems. We limit the size of the reservoir to emulate a scenario in which large reservoirs are undesirable due to computational constraints (as is typical for harder prediction problems). In typical traditional RC implementations, computational costs scale as poorly as $O(N^3)$, where $N$ is the number of neurons in the reservoir. Hence, methods that can reduce the size of the reservoir while maintaining accuracy are highly valuable for many applications. We also consider the common scenarios in which the time series data cannot be sampled at small intervals due to costs or feasibility of data collection and/or the NGRC feature vector doesn't contain all the necessary nonlinearities in the system. We show that in cases where the NGRC is limited, a hybrid RC-NGRC approach using a small reservoir component reaches the performance level of a large traditional RC, while offering much greater computational efficiency. 

In Section \ref{sec:RC_and_NGRC}, we review the use of RCs and NGRCs for time series prediction. In Section \ref{sec:hybrid}, we present the details of our hybrid RC-NGRC approach. We show results applying this approach to forecasting the prototypical Lorenz system and other chaotic dynamical systems in Section \ref{sec:results}. We conclude in Section \ref{sec:conclusion} that a hybrid RC-NGRC approach is particularly useful in situations where computational resources are limited and standalone NGRCs struggle.

\section{Background: Traditional and next-generation reservoir computing}
\label{sec:RC_and_NGRC}

Suppose we have discrete time series data $\{ \vb{u}(\tau), \vb{u}(2 \tau), \ldots \}$ sampled at regular time steps from the trajectory $\vb{u}(t) \in \mathbb{R}^d$ of a $d$-dimensional dynamical system. Using the first $n_{\text{train}}$ data points as training data, the goal of RC and NGRC forecasting is to produce a predicted trajectory $\vb{v}(t)$ for time $t > t_{\text{train}}$ (where $t_{\text{train}} = n_{\text{train}} \tau$) that is a good match to $\vb{u}(t)$. In addition to seeking a high quality short term forecast with $\vb{v}(t) \approx \vb{u}(t)$ for as long as possible after $t_{\text{train}}$, we also seek to replicate the system's climate, meaning that the long term statistical features of $\vb{v}$ match those of  $\vb{u}$.

\subsection{Reservoir computers (RCs)}
\label{sec:RC}

A typical RC (Figure \ref{fig:schematics}a)) uses a random artificial  neural network with recurrent links of fixed weights as a reservoir. To define such a reservoir, we initialize a random directed unweighted network of $N$ nodes with a Poisson degree distribution having average degree $\langle k \rangle$. We multiply the nonzero elements of the resulting Boolean adjacency matrix $\tilde{A} \in \mathbb{R}^{N \times N}$ by link weights chosen from the uniform distribution on $[-1, 1]$ to form the matrix $A$, and then we rescale $A$ to have spectral radius $\rho$. Typically, $\rho \lessapprox 1$, though for some problems $\rho$ near zero can be appropriate \cite{griffith.etal_2019, jaurigue_2024}; further discussion and investigation can be found in the Supplementary Materials.

At all time steps of the training data, the state of the reservoir $\vb{r}(t) \in \mathbb{R}^N$ depends on the input it receives, $\vb{u}(t)$, and its state at the previous time step, $\vb{r}(t-\tau)$, through the relationship
\begin{equation} \label{eqn:RC_update}
    \vb{r}(t) = (1 - \alpha) \vb{r}(t - \tau) + \alpha f(A \vb{r}(t - \tau) + B \vb{u}(t) + c),
\end{equation}
where $f$ is the hyperbolic tangent function (applied element-wise), and we choose the entries of the input matrix $B \in \mathbb{R}^{N \times d}$ from the uniform distribution on $[-\sigma, \sigma]$; $\alpha$ is a leakage parameter that controls the timescale of the reservoir's response to its input.

To train the RC, we first synchronize the reservoir by initializing it in the zero state ($\vb{r}(0) = 0$) and feeding in the first $n_{\text{warmup}}$ data points (up to time $t_{\text{warmup}} = n_{\text{warmup}} \tau$) according to Equation \ref{eqn:RC_update}. We then feed in the remaining $n_{\text{fit}} = n_{\text{train}} - n_{\text{warmup}}$ training data points (spanning a time $t_{\text{fit}} = n_{\text{fit}} \tau$), and train a readout matrix $W \in \mathbb{R}^{d \times N}$ to make a one-step-ahead prediction for this data:
\begin{equation} \label{eqn:RC_W_approx}
    W \vb{r}(t) \approx \vb{u}(t + \tau), \quad t_{\text{warmup}} < t < t_{\text{train}}.
\end{equation}
We fit $W$ using ridge regression (linear regression with Tikhonov regularization), which minimizes the quantity
\begin{equation} \label{eqn:tikhonov_loss}
    \sum_{t_{\text{warmup}} < t < t_{\text{train}}} 
    \left( \norm{W \vb{r}(t) - \vb{u}(t + \tau)}^2 \right) + \beta \Tr(W W^T).
\end{equation}
Here, $\beta$ is the regularization hyperparameter that penalizes large entries of $W$ to prevent overfitting.

We note that in practice, we use the input noise technique of adding weak Gaussian noise (standard deviation $\gamma \ll 1$) to $\vb{u}$ before feeding it in to the reservoir via Equation \ref{eqn:RC_update}, but using noiseless data to fit $W$, as this has been shown to promote climate stability of autonomous predictions \cite{wikner.etal_2024} (see the Supplementary Materials).

After training the readout matrix $W$, we switch the reservoir to autonomous mode (Figure \ref{fig:schematics}a)) in which the output of the reservoir $W \vb{r}(t)$ is repeatedly fed back in as input:
\begin{align} \label{eqn:RC_closed_loop}
    &\vb{v}(t) = W \vb{r}(t - \tau) \\
    &\vb{r}(t) = (1 - \alpha) \vb{r}(t - \tau) + \alpha f(A \vb{r}(t - \tau) + B \vb{v}(t) + \vb{c})
\end{align}
for $t = t_{\text{train}} + \tau, t_{\text{train}} + 2\tau, \ldots$. The autonomous forecast for the trajectory of the dynamical system is then $\vb{v}(t)$.

Reservoir computing offers a state-of-the-art method for time series forecasting of dynamical systems, capable of both short term forecast accuracy and long term climate replication. Compared to deep neural networks, RCs have dramatically reduced training time because only the output weights are fit \cite{vlachas.etal_2020}. However, training and/or simulating a large reservoir can still be computationally expensive in some cases: computational costs of fitting an output matrix scale as poorly as $\mathcal{O}(N^3)$ in typical RC implementations, and well-performing RCs may require large $N$ on the order of $1000$ or more for low-dimensional dynamical systems and much more for higher dimensional systems. Furthermore, there are many hyperparameters (listed in Table \ref{tab:hybrid_hyperparameters}) that must be tuned for optimal or near-optimal performance, making RCs nontrival to implement in many cases.

\subsection{Next-Generation Reservoir Computers (NGRCs)}
\label{sec:NGRC}

In contrast to RCs, NGRCs utilize a feature vector constructed directly from the training data $\vb{u}(t)$ in order to make predictions (Figure \ref{fig:schematics}b)). First, we must specify the number $k$ of current and time-delayed observations in the feature vector, and the number $s$ of time steps between successive time-delayed observations. In our studies we focus on $k=2$ and $s=1$. Then, at each time step of the training data we construct the feature vector $\vb{O}(t)$ as follows:
\begin{itemize}
    \item We concatenate the current and time-delayed observations to form a linear feature vector:
    \begin{equation} \label{eqn:NGRC_linear_feature_vector}
        \vb{O}_{\text{lin}}(t) = \vb{u}(t) \oplus \vb{u}(t - s\tau) \oplus \ldots \oplus \vb{u}(t - (k-1) s\tau),
    \end{equation}
    where $\oplus$ represents vector concatenation.
    As in the RC case, in practice we add weak Gaussian noise with standard deviation $\gamma \ll 1$ to $\vb{u}$ before forming $\vb{O}_{\text{lin}}$ (see the Supplementary Materials).
    \item We form a nonlinear feature vector $\vb{O}_{\text{nonlin}}(t)$ consisting of nonlinear functions of the elements of $\vb{O}_{\text{lin}}(t)$. Here, we choose to form $\vb{O}_{\text{nonlin}}(t)$ by listing all unique quadratic monomials of the linear terms, e.g. $u_2(t)u_1(t - s\tau)$.
    \item We form the full feature vector by concatenating a constant element (taken to be 1), the linear feature vector, and the nonlinear feature vector: 
    \begin{equation} \label{eqn:NGRC_feature_vector}
        \vb{O}(t) = 1 \oplus \vb{O}_{\text{lin}}(t) \oplus \vb{O}_{\text{nonlin}}(t)
    \end{equation}
\end{itemize}
Note that $\vb{O}(t)$ is defined for $s (k-1) \tau < t \leq t_{\text{train}}$; as such, $s (k-1) \tau$ is the effective warm up time of the NGRC. At each time step, there are $1 + dk + dk (dk+1) / 2$ elements of the feature vector $\vb{O}(t)$, where $d$ is the dimension of $\vb{u}$ (e.g. $28$ elements for $k=2, d=3$). 

Next, $\vb{O}(t)$ plays an analogous role to $\vb{r}(t)$ in an RC. We fit a readout matrix $W$ using ridge regression (see Equation \ref{eqn:tikhonov_loss}) to satisfy
\begin{equation} \label{eqn:NGRC_W_approx}
    W \vb{O}(t) \approx \vb{u}(t + \tau), \quad s (k-1) \tau < t < t_{\text{train}}.
\end{equation}
Then, we use autonomous mode to make a prediction for $t = t_{\text{train}} + \tau, t_{\text{train}} + 2\tau, \ldots$:
\begin{itemize}
    \item We make a one-step prediction: $\vb{v}(t) = W \vb{O}(t - \tau)$.
    \item We construct $\vb{O}(t)$ according to Equation \ref{eqn:NGRC_feature_vector}.
    In doing so, we draw time-delayed terms from $t_{\text{train}}$ and before from $\vb{u}$, and draw those from after $t_{\text{train}}$ from $\vb{v}$.
\end{itemize}
The NGRC forecast for the trajectory of the dynamical system is then $\vb{v}(t)$.

NGRCs have strong predictive ability and have some important advantages: compared to RCs, NGRCs are more computationally efficient due to having many fewer terms in their feature vector, require less hyperparameter tuning, and have a very small effective warm up time of $s (k-1) \tau$ \cite{gauthier.etal_2021}. However, NGRCs have their own drawbacks: performance can be very dependent on the choice of nonlinear functions used to construct $\vb{O}_{\text{nonlin}}$, in some cases showing poor performance if the specific nonlinearities of the true dynamical system are not reflected in the feature vector \cite{zhang.cornelius_2023}. We will also show later in Section \ref{sec:hybrid_improvement_large_timestep} that NGRCs struggle when the training data are sampled sparsely from the true system, i.e. the time step $\tau$ is large.

\section{Methods: Hybrid RC-NGRC forecasting approach}
\label{sec:hybrid}

We now introduce the central innovation of our paper: a hybrid RC-NGRC scheme for forecasting dynamical systems. In the hybrid RC-NGRC scheme, we utilize both a small reservoir and an NGRC feature vector to make a forecast (Figure \ref{fig:schematics}c)). First, we initialize a reservoir just as in Section \ref{sec:RC}. In practice we usually use a lightweight reservoir with a small number of nodes $N \leq 100$. Then, we use training data to construct both a reservoir state $\vb{r}(t)$ and an NGRC feature vector $\vb{O}(t)$ for all possible time steps, as specified in Equations \ref{eqn:RC_update} and \ref{eqn:NGRC_linear_feature_vector}, \ref{eqn:NGRC_feature_vector}. We form a hybrid feature vector
\begin{equation} \label{eqn:hybrid_representation}
    \vb{H}(t) = \vb{r}(t) \oplus \vb{O}(t)
\end{equation}
at each time step, where $\oplus$ represents concatenation of vectors. Then we fit $W$ using ridge regression (see Equation \ref{eqn:tikhonov_loss}) to best satisfy
\begin{equation} \label{eqn:hybrid_W_approx}
    W \vb{H}(t) \approx \vb{u}(t + \tau), \quad t_{\text{warmup}} < t < t_{\text{train}}.
\end{equation}
After training the readout matrix $W$, we produce an autonomous prediction $\vb{v}$ by iterating the equations below:
\begin{align} \label{eqn:hybrid_closed_loop}
    &\vb{v}(t) = W \vb{H}(t - \tau) \\
    &\vb{r}(t) = (1 - \alpha) \vb{r}(t - \tau) + \alpha f(A \vb{r}(t - \tau) + B \vb{v}(t) + \vb{c}) \\
    &\vb{O}(t) = 1 \oplus \vb{O}_{\text{lin}}(t) \oplus \vb{O}_{\text{nonlin}}(t) \\
    &\vb{H}(t) = \vb{r}(t) \oplus \vb{O}(t).
\end{align}
(see Section \ref{sec:NGRC} for more detail on constructing $\vb{O}$).

\begin{table}
\caption{\label{tab:hybrid_hyperparameters} Details of the training data and hyperparameters of the RCs and NGRCs used to make forecasts. The hybrid RC-NGRC uses the same hyperparameters. These values are used throughout, except where otherwise noted.}
\begin{tabular}{lll}
\hline
\multirow{3}{*}{\begin{tabular}[c]{@{}l@{}}Training\\ data\end{tabular}} & Time step                                                                              & $\tau = 0.06$               \\
                         & \# training data points                                                                       & $n_{\text{train}} = 10,000$             \\
                          & Noise standard deviation                                                                            & $\gamma = 1 \times 10^{-3}$ \\ \hline
\multirow{9}{*}{Reservoir}     & Number of nodes                                                                           & $N = 50$                   \\
                               & Average degree                                                                            & $\langle k \rangle = 10$    \\
                               & Spectral radius                                                                           & $\rho = 0.9$               \\
                               & Leakage rate                                                                              & $\alpha = 1$                \\
                               & Bias                                                                                      & $c = 0.5$                     \\
                               & Input matrix scaling                                                                      & $\sigma = 1$                \\
                               & Warm-up time steps                                                                        & $n_{\text{warmup}} = 1000$ \\
                               & Regularization parameter                                                                  & $\beta = 1 \times 10^{-8}$  \\ \hline
\multirow{3}{*}{NGRC}          & \begin{tabular}[c]{@{}l@{}}Number of current and\\ time-delayed observations\end{tabular} & $k = 2$                     \\
                               & \begin{tabular}[c]{@{}l@{}}Spacing of time-\\ delayed observations\end{tabular}           & $s = 1$                     \\
                               & Regularization parameter                                                                  & $\beta = 1 \times 10^{-8}$  \\ \hline
\end{tabular}
\end{table}

In practice, to create hybrid RC-NGRC, RC, and NGRC predictions of a given chaotic dynamical system with an attractor, we use the following procedure. We sample a random initial condition from the attractor, integrate the system forward using the fourth order Runge-Kutta method with an integration time step $\tau_{\text{int}} = 0.001 \ll \tau$, and subsample with time step $\tau$ to obtain the time series data $\vb{u}(t)$. We normalize $\vb{u}$ so that each component of the training data (first $n_{\text{train}}$ data points) has mean $0$ and standard deviation $1$. We generate a random realization of the reservoir as described in Section \ref{sec:RC}, using the hyperparameters given in Table \ref{tab:hybrid_hyperparameters}. Then, we construct autonomous predictions of length $t_{\text{predict}}$ using the RC, NGRC, and hybrid RC-NGRC (Sections \ref{sec:RC}, \ref{sec:NGRC}, and \ref{sec:hybrid}), using the input noise technique (see Supplementary Materials). For multiple trials, we repeat the whole procedure above, so that different trials have both different initial conditions of the trajectory and different random reservoir realizations.

\section{Results}
\label{sec:results}

\begin{figure*}
    \includegraphics[width=\textwidth]{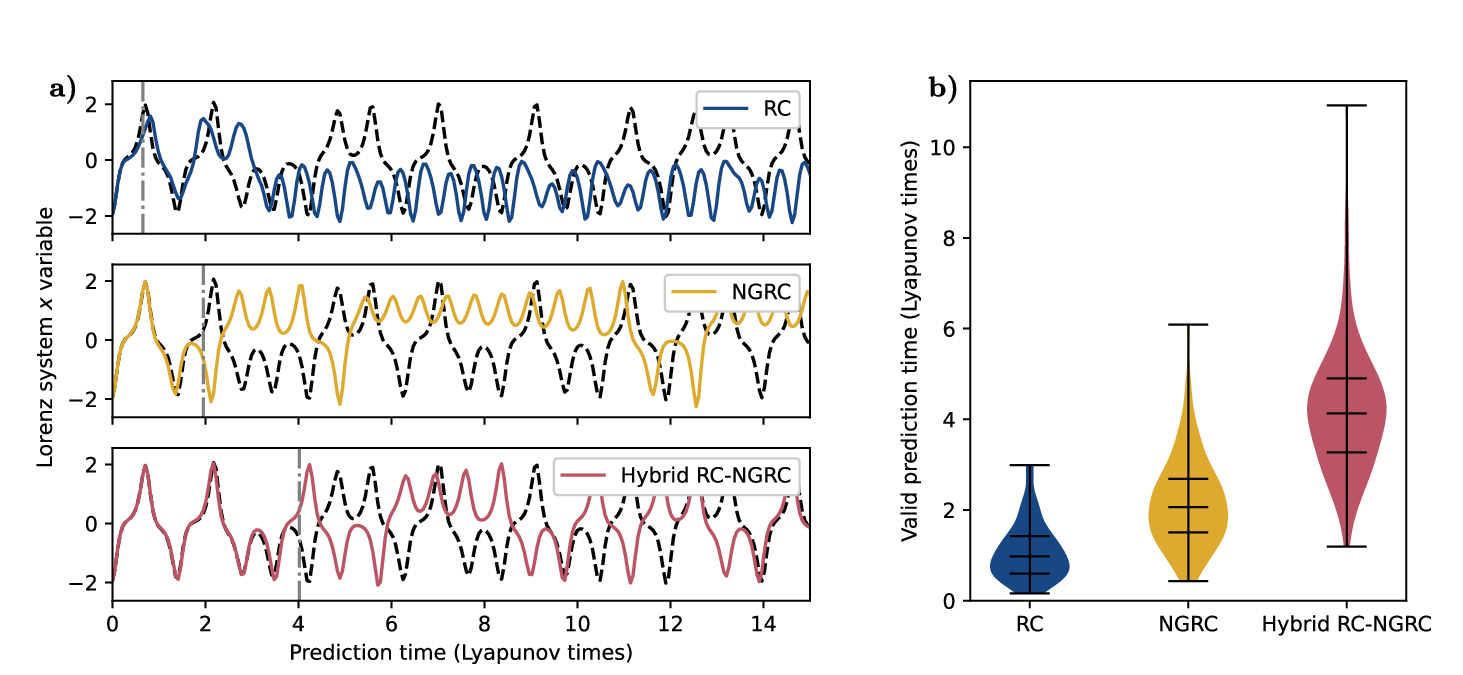}
    \caption{\label{fig:hybrid_VPTs} \textbf{a)} Representative examples of RC, NGRC, and hybrid RC-NGRC autonomous predictions of the Lorenz system ($x$ component shown), where a small reservoir ($N=50$) and large time step ($\tau = 0.06$) are used, limiting RC and NGRC performance. Valid prediction time (VPT) indicated by the vertical dashed line. \textbf{b)} Distributions of VPTs for RC, NGRC, and RC-NGRC predictions, where each trial is done on new initial conditions using a new reservoir realization. The hybrid RC-NGRC shows substantially stronger short term predictive power than either the RC or NGRC alone. Horizontal lines: quartiles (100 trials).}
\end{figure*}

\subsection{Forecasting the Lorenz system with a small reservoir and large time step}
\label{sec:hybrid_lorenz}

We now evaluate the hybrid RC-NGRC approach on the task of predicting the Lorenz system, a prototypical chaotic dynamical system governed by the equations
\begin{equation} \label{eqn:lorenz}
    \dot{x} = 10 (y - x), \:
    \dot{y} = x (28 - z) - y, \:
    \dot{z} = x y - 8 z / 3 
\end{equation}
\cite{lorenz_1963}. We compare to the standalone RC and NGRC components, with hyperparameters as listed in Table \ref{tab:hybrid_hyperparameters}. Although both RC and NGRC approaches are capable of forecasting the Lorenz system under ideal conditions, here we impose additional constraints on the prediction methods. We limit the size of the reservoir, imitating a scenario in which large reservoirs are not desirable due to computational constraints, and we use training data that is sampled from the Lorenz system at a large time step, imitating a scenario in which observing the state of the dynamical system can only be done sparsely. These constraints make the task of forecasting the Lorenz system formidable for both RCs and NGRCs.

\subsubsection{Short term forecast quality}
\label{sec:hybrid_lorenz_short_term}

To evaluate the quality of a prediction in the short term, we use valid prediction time (VPT) as a metric, defined as the time at which the root mean square error of the normalized prediction exceeds a threshold, here chosen to be $\kappa = 0.9$:
\begin{equation} \label{eqn:VPT_definition}
    t_{\text{VPT}} = \min{ \left \{ t \, : \, \frac{\norm{\vb{v}(t) - \vb{u}(t)}}{\sqrt{\langle \norm{\vb{u}(t)}^2 \rangle}} > \kappa \right \} } - t_{\text{train}}.
\end{equation}
In chaotic systems such as the Lorenz system, errors grow approximately as $\exp(\Lambda_{\text{max}} t)$ where $\Lambda_{\text{max}}$ is the maximal Lyapunov exponent. Thus, the Lyapunov time $t_{\text{lyap}} = \Lambda_{\text{max}}^{-1}$ is a natural timescale for evaluating the quality of forecast of a chaotic dynamical system, and we report $t_{\text{VPT}}$ in units of $t_{\text{lyap}}$.

Representative examples of predictions $\vb{v}(t)$ of the Lorenz system by RC, NGRC, and hybrid RC-NGRC prediction schemes are shown in Figure \ref{fig:hybrid_VPTs}a) with VPTs marked (only the $x$ coordinates of the predictions are shown). In this example, the VPT of the hybrid RC-NGRC forecast is much longer than for the RC or NGRC alone. Recording VPTs over many trials with different Lorenz system initial conditions and different random reservoirs yields a distribution of VPTs shown in Figure \ref{fig:hybrid_VPTs}b). We find that hybrid RC-NGRC has much better short term predictive power than either the RC or NGRC, as evidenced by the longer median valid prediction times (VPTs) (hybrid: $4.13 t_{\text{lyap}}$, RC: $0.98 t_{\text{lyap}}$, NGRC: $2.06 t_{\text{lyap}}$). We observe similar results without input noise (see Supplementary Materials).

\begin{figure*}
    \includegraphics[width=\textwidth]{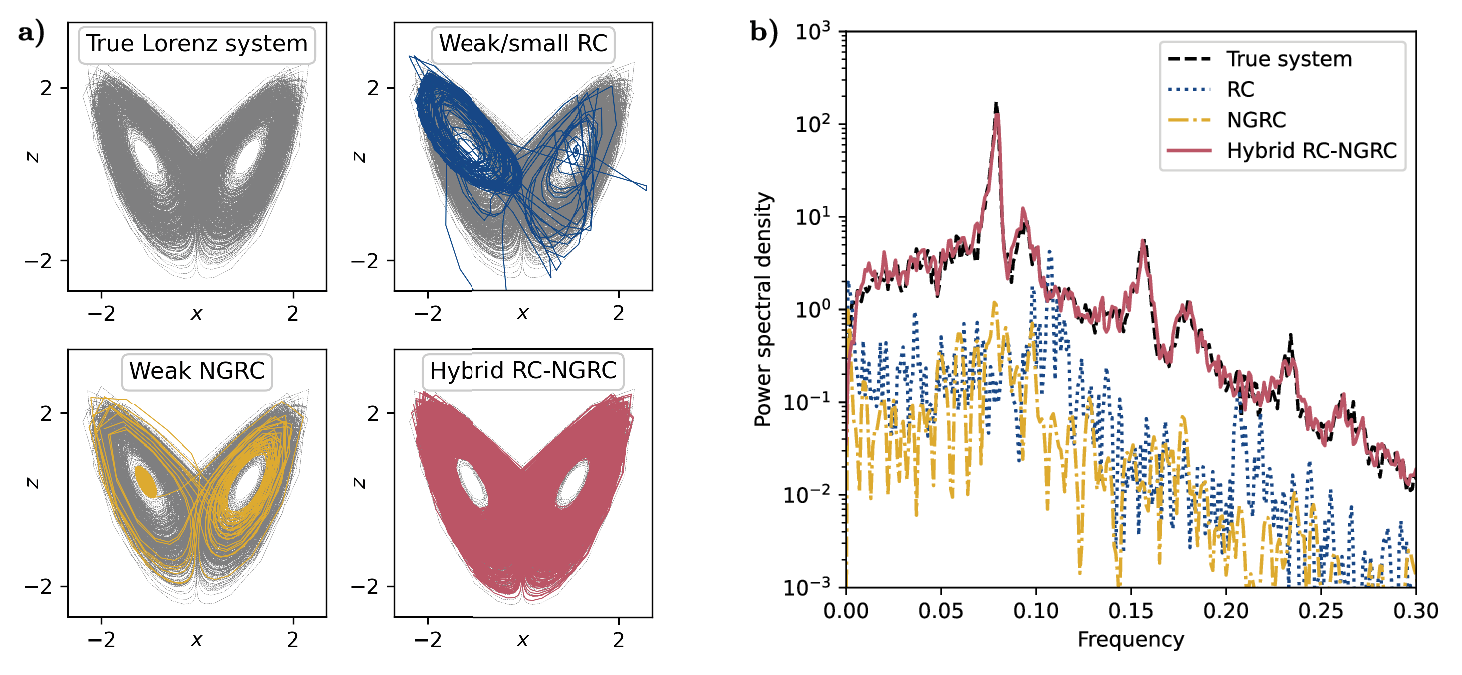}
    \caption{\label{fig:hybrid_climates} \textbf{a)} Representative examples of long-term phase space trajectories of the RC, NGRC, and hybrid RC-NGRC autonomous predictions (predictions extended from Figure \ref{fig:hybrid_VPTs}a)). Though RC and NGRC reconstruct the Lorenz attractor in some trials, only the hybrid RC-NGRC prediction reliably reconstructs the attractor of the true system across trials. \textbf{b)} Power spectra of $z$ component of autonomous predictions for the different methods. Only the hybrid RC-NGRC prediction reliably reproduces the spectrum of the Lorenz system.}
\end{figure*}

\subsubsection{Long term climate replication}
\label{sec:hybrid_lorenz_long_term}

We also find that the hybrid RC-NGRC more accurately reproduce the climate (long-term statistical properties) of the Lorenz system when compared to either the RC or NGRC alone. Two-dimensional projections of the phase space trajectories of the representative predictions from Figure \ref{fig:hybrid_VPTs}a) are plotted in Figure \ref{fig:hybrid_climates}a). In this example, the RC forecast fails to recreate the butterfly-shaped attractor. The NGRC prediction initially tracks the true attractor, but then gets trapped orbiting inward toward one of the unstable fixed points of the true system. Although in some trials the RC and NGRC predictions can recreate the true attractor, they often fail in a similar manner as in Figure \ref{fig:hybrid_climates}a). The climate reproduction of especially the NGRC is even worse when not using the input noise technique (see the Supplementary Materials). In contrast, the hybrid RC-NGRC approach robustly succeeds at accurately reconstructing the butterfly-shaped strange attractor of the Lorenz system.

We also examine the power spectral densities of the different methods' forecasts to determine whether they recover the climate of the Lorenz system. Figure \ref{fig:hybrid_climates}b) shows the power spectra of the $z$ components of the same representative predictions, found using Welch's method \cite{welch_1967}. The power spectrum of the hybrid RC-NGRC prediction matches that of the Lorenz system nearly perfectly, suggesting the prediction captures the statistical properties of the Lorenz system. In contrast, the spectra of the RC and NGRC predictions fail to capture the features of the true system's spectrum.

\begin{table*}
    \caption{\label{tab:map_errs} Quantifying the error in climate replication: mean normalized map errors $\overline{\epsilon}_{\text{map}}$ of forecasts with standard errors, averaged over 64 trials. The lowest value (best performing) for each system is bolded.}
    \begin{tabular}{@{}llll@{}}
    \hline
                                                                                                            & RC \qquad & NGRC \quad & \begin{tabular}[c]{@{}l@{}}Hybrid \\ RC-NGRC \end{tabular} \\ \hline
    Lorenz & $(8.8 \pm 1.5) \times 10^{-1}$ \qquad \qquad & $(3.9 \pm 0.2) \times 10^{-2}$  \qquad \qquad & $\mathbf{ (6.0 \pm 0.1) \times 10^{-3} }$            \\ 
    R\"ossler & $(2.8 \pm 1.5) \times 10^{0}$ & $(1.5 \pm 0.1) \times 10^{-2}$  & $\mathbf{ (3.7 \pm 0.2) \times 10^{-3} }$            \\ 
    Double Scroll \qquad \qquad & $(6.2 \pm 1.7) \times 10^{-1}$ & $(2.0 \pm 0.4) \times 10^{0}$  & $\mathbf{ (8.2 \pm 1.2) \times 10^{-2} }$            \\ \hline
    \end{tabular}
\end{table*}

To quantify the error in climate replication, we use the normalized map error $\epsilon_{\text{map}}(t)$ introduced in Ref. \cite{wikner.etal_2024}, which for a forecast $\vb{v}$ at time $t$ is defined as
\begin{equation} \label{eqn:map_err_defn}
    \epsilon_{\text{map}}(t) = \frac{\norm{\vb{v}(t) - \vb{F}(\vb{v}(t - \tau), \tau)}}{\overline{E}_{\text{map}}}
\end{equation}
where $\vb{F}(\vb{v_0}, \tau)$ is a function that integrates the true evolution equations forward from initial condition $\vb{v_0}$ for time $\tau$, in practice using the integration time step $\tau_{\text{int}} \ll \tau$. The normalization constant is the mean error of the persistence forecast of the training data, $\overline{E}_{\text{map}} = \overline{\norm{\vb{u}(t + \tau) - \vb{u}(t)}}$, where the mean is taken over all training data. The mean normalized map error $\overline{\epsilon}_{\text{map}}$, the mean of $\epsilon_{\text{map}}(t)$ over the first $n_{\text{predict}}$ time steps, quantifies how faithful single steps of the prediction $\vb{v}(t)$ are to the system's true equations on average. We emphasize that this is different than just calculating the open-loop one-step test error, because it takes into account the regions of phase space explored by the autonomous closed-loop system that we employ for multi-step predictions. We also note that the map error can only be calculated for systems in which the true equations are known, such as the model systems studied here. The results in the first row of Table \ref{tab:map_errs} show that the mean normalized map errors of hybrid RC-NGRC forecasts of the Lorenz system are substantially lower than those of RC or NGRC forecasts. Our results on the Lorenz system suggest that in cases when both the RC and NGRC are limited, e.g. for a small reservoir and large sampling time step in the training data, the hybrid RC-NGRC method offers a significant improvement in climate replication over either the RC or NGRC alone.

\subsection{Under what conditions is the hybrid RC-NGRC approach particularly advantageous?}
\label{sec:hybrid_improvement}

Here, we relax the constraints from the previous section that the reservoir be small and the training data be sampled with large time step. We find that although the hybrid RC-NGRC still achieves good predictive performance, it loses its relative advantage over RC and/or NGRC approaches. This occurs as we enter a regime where either the RC or NGRC perform very well, thus eliminating the need for a hybrid approach. Our results suggest that the greatest utility of the hybrid RC-NGRC comes when the NGRC is limited (e.g. because the training data is sampled at a large time step) and computational efficiency is a priority (making small reservoirs highly advantageous).

\subsubsection{Weak RC}
\label{sec:hybrid_improvement_small_RC}

\begin{figure*}
    \includegraphics[width=\textwidth]{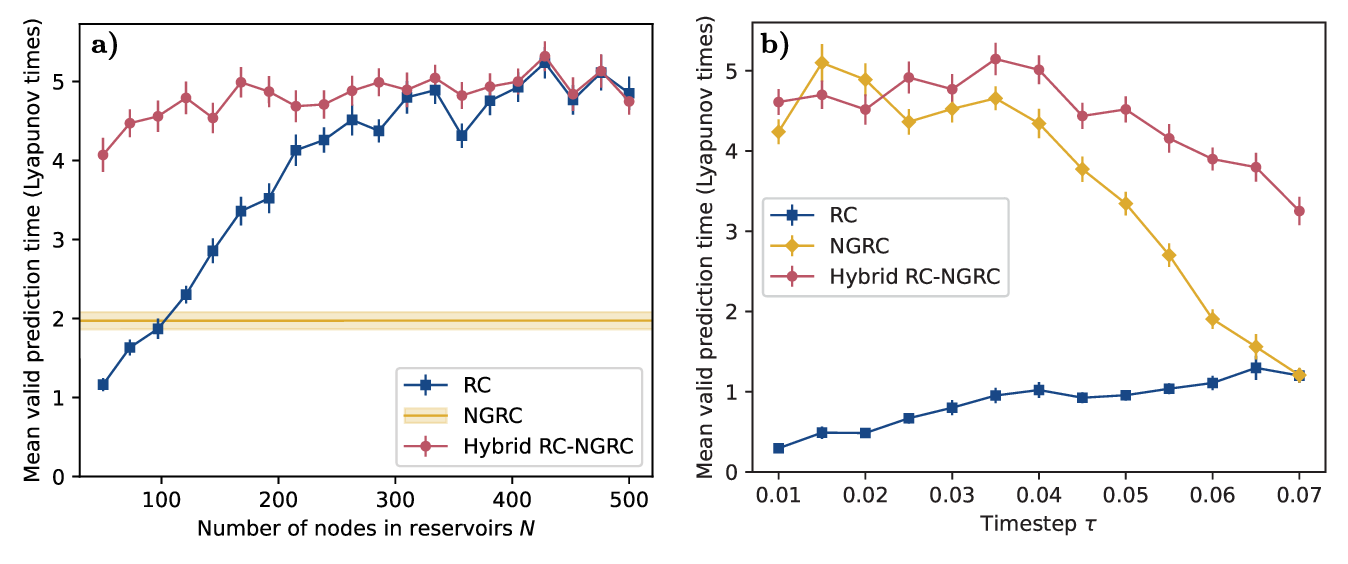}
    \caption{\label{fig:VPT_vs_size_and_timestep} \textbf{a)} Mean valid prediction times for the Lorenz system versus number of nodes in the reservoir. Although RC performance is poor at small $N$, and NGRC performance is modest due to using a large timestep ($\tau = 0.06$), the hybrid RC-NGRC performs well throughout, specifically providing a substantial advantage over both RC and NGRC at small $N$. Note that the hybrid RC-NGRC approach with reservoir size N = 100 approximately matches that of a pure RC with N = 500. \textbf{b)} Mean valid prediction times for the Lorenz system versus time step size $\tau$ in the training data. As time step is adjusted, the number of training data points $n_{\text{train}}$ is kept constant. The hybrid RC-NGRC shows the greatest advantage in predictive power over the RC or NGRC alone when using a large time step. Reservoir size $N = 50$. Error bars and band: standard error of the mean (64 trials).}
\end{figure*}

The predictive power of an RC is predicated on having a high-dimensional reservoir with enough fitting parameters to accurately capture the behavior of the dynamical system. In practice, as the number of reservoir nodes $N$ is increased, the forecasting skill of the RC increases, until some saturation point, as reflected in Figure \ref{fig:VPT_vs_size_and_timestep}a). However, Figure \ref{fig:VPT_vs_size_and_timestep}a) also shows that the hybrid RC-NGRC forecasts achieve a similar mean valid prediction time using a much smaller reservoir (compare hybrids with 100 nodes to RCs with 500 nodes). This is true even though the NGRC itself is poorly performing (due to large time step). Hybridizing an RC with even a poorly performing NGRC enables strong predictive ability even with a very small reservoir.

\begin{figure}
    \includegraphics[width=0.5\textwidth]{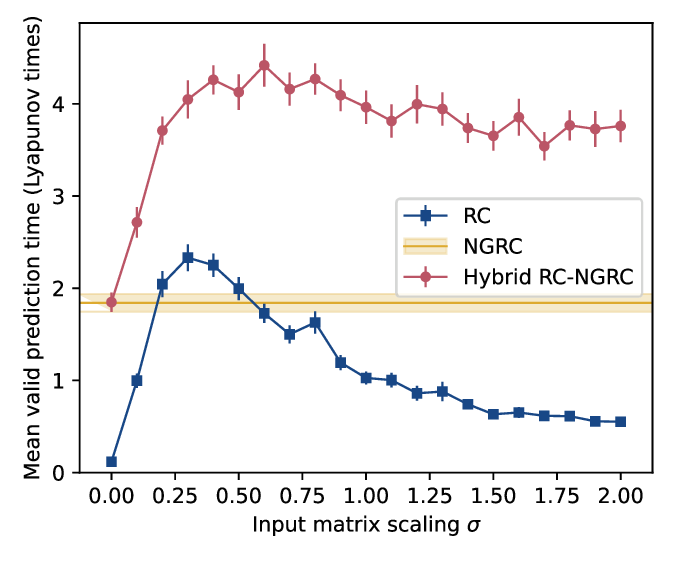}
    \caption{\label{fig:VPT_vs_sigma} The hybrid RC-NGRC approach exhibits reduced sensitivity to reservoir hyperparameters compared to RCs. Shown here: mean VPT vs. input matrix scaling $\sigma$ (other examples shown in Supplementary Materials). Error bars and band: standard error of the mean (64 trials).}
\end{figure}

We also observe that the hybrid RC-NGRC exhibits substantially reduced sensitivity to reservoir hyperparameters compared to the RC alone. Figure \ref{fig:VPT_vs_sigma} shows mean VPT vs. one reservoir hyperparameter, the input matrix scaling $\sigma$ (more examples are shown in the Supplementary Materials). The hybrid RC-NGRC functions well, maintaining a nearly saturated mean VPT for a large range of the hyperparameters, even when its RC component is clearly sub-optimal. These suggests that, as compared with traditional RCs, the hybrid method allows the user to avoid careful hyperparameter tuning, providing additional computational advantages beyond those gained by reducing the reservoir size. 

\subsubsection{Weak NGRC}
\label{sec:hybrid_improvement_large_timestep}

While NGRCs have been shown to work in a range of cases, their most accurate predictions are achieved when the specific nonlinearities of the underlying system appear in the NGRC represenation vector \cite{zhang.cornelius_2023}. In this case, during training the NGRC can learn weights that make the autonomous mode imitate a numerical integrator of the true system. However, just as numerical integration methods can fail if the integration time step is too large, the NGRC can also fail if the time step is too large. At large time steps, the output after a single time step is not well-approximated by the difference-equation approximation of the true equations governing the system. In Figure \ref{fig:VPT_vs_size_and_timestep}b), we plot the valid prediction times for the NGRC for the Lorenz system as as a function of time step length $\tau$. In the case of the Lorenz system, the NGRC feature vector we're using contains all the contributing nonlinearities at small time steps. Note that as $\tau$ is varied, the number of training data time steps $n_{\text{train}}$ is kept constant, so for larger $\tau$, $t_{\text{train}} = n_{\text{train}} \tau$ is greater. We see that the valid prediction time of NGRC predictions decrease as $\tau$ is increased. We see a modest increase in the VPT for the RC as $\tau$ is increased, because the number of training data time steps is kept constant, the information content in the training signal initially increases as $\tau$ increases from a small value. 

Compared with the NGRC, the drop off of the hybrid RC-NGRC's valid prediction time as time step is increased is much less dramatic. Although the RC alone shows weak predictive power for all sampling time steps, at large time steps the hybrid's valid prediction time is much greater than either the RC or NGRC alone. When using data for which NGRC performance is poor due to large sampling time step, hybridizing even a poorly performing RC with an NGRC can dramatically improve the prediction performance. 

In Section \ref{sec:hybrid_other} we show that for another chaotic system (Double Scroll) where the NGRC offers weak performance because its feature vector does not contain the essential nonlinearities, we see a similar performance advantages using the hybrid approach with a small RC. However, these advantages do not depend on the sampling time step, as the NGRC's performance remains weak at small time steps.

\subsection{How performance depends on the size of the training dataset}
\label{sec:training_data_requirements}

\begin{figure}
    \includegraphics[width=0.5\textwidth]{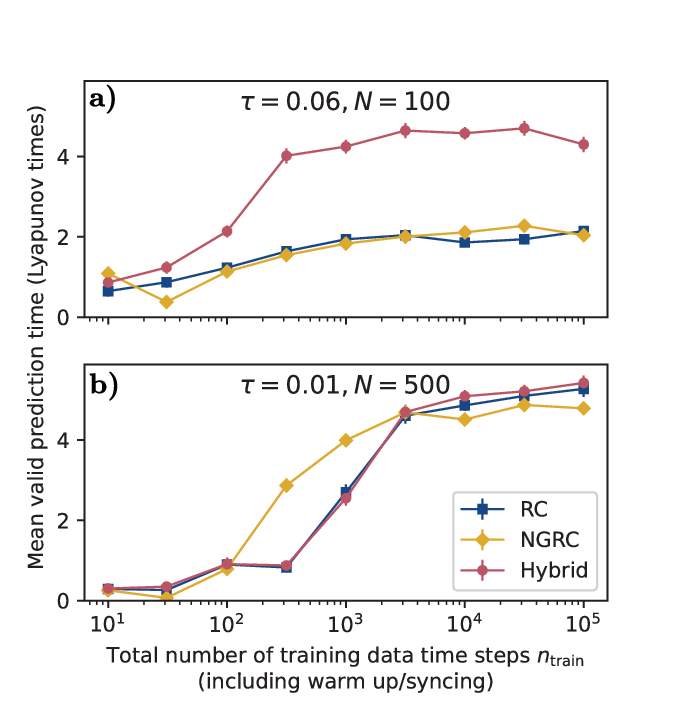}
    \caption{\label{fig:VPT_vs_training_data} Mean valid prediction times for the Lorenz system versus number of training data points, in scenarios where \textbf{a)} RC and NGRC struggle, and \textbf{b)} RC and NGRC perform well with enough training data. Regularization strength is scaled proportionally to $n_{\text{train}}$. Error bars (where visible): standard error of the mean (64 trials).}
\end{figure}

NGRCs are touted as requiring shorter training datasets than RCs, owing to their very short effective warm up time of $s (k-1) \tau$ (just a single time step for $k=2$, $s=1$) \cite{gauthier.etal_2021}. In contrast, RCs must have enough warm up time $t_{\text{warmup}}$ at the beginning of training to synchronize the reservoir to the input data. For RCs whose internal state reflects low memory of prior inputs (such as those with small $\rho$ investigated in the Supplementary Materials) only a short $t_{\text{warmup}}$ is needed, but for typical RC implementations with longer memories, the long $t_{\text{warmup}}$ can be an obstacle, especially if training data availability is limited. Here we explore how a hybrid RC-NGRC, which also requires a warm up time for its reservoir, performs when training on a limited number of time steps.

We vary the total number of time steps $n_{\text{train}}$ of Lorenz system training data supplied to the RC, NGRC, and hybrid RC-NGRC, and plot the mean VPTs of each model. We perform this analysis a) with a relatively small reservoir and large time step ($\tau = 0.06$, $N = 100$), and b) in an easier scenario with large reservoir and small time step ($\tau = 0.01$, $N = 500$). As we vary $n_{\text{train}}$, we also scale the regularization hyperparameter $\beta$ accordingly: $\beta  = n_{\text{train}} \tilde{\beta}$ with $\tilde{\beta} = 1 \times 10^{-12}$. In all trials, to choose a warm up time for the RC, we first initialize two copies of the reservoir that are identical except for their initial reservoir states $\vb{r^{(1)}}(0) \neq \vb{r^{(2)}}(0)$, and feed the same input data into both reservoirs. We take the inverse slope of a linear fit of $\ln \left( |\vb{r^{(1)}}(t) - \vb{r^{(2)}}(t)| \right)$ vs. $t$ as the empirical characteristic time $t_{\text{sync}}$ for reservoir synchronization, then use a warm up time $t_{\text{warmup}} = 10 t_{\text{sync}}$ (capped to a maximum of $t_{\text{train}} / 4$). After $t_{\text{warmup}}$ the synchronization error in the reservoir state will be on the order of $e^{-10} \approx 5 \times 10^{-5}$. In practice, we find $t_{\text{warmup}} \approx 20 \tau$ for these hyperparameters. (Note that if we used smaller leakage rates $\alpha$, we would expect the warm up time to be longer.) For the hybrid RC-NGRC, we use the same $t_{\text{warmup}}$ as for RC. For the NGRC, the effective warm up length is always only $s (k-1) \tau$.

In the scenario where both RC and NGRC struggle due to small reservoir and large time step (Figure \ref{fig:VPT_vs_training_data}a)), we find that the hybrid RC-NGRC maintains it relative advantage over NGRCs and RCs over a wide range of training data amounts. Despite the NGRC having a much shorter warm up time compared to the hybrid, using the hybrid RC-NGRC yields improved performance even when using training data amounts down to $\sim 100$ steps. 

However, when a well-performing NGRC is available, the hybrid RC-NGRC loses out to NGRC in the low training data regime. In Figure \ref{fig:VPT_vs_training_data}b), we plot VPT versus training data amount in the scenario where both RC and NGRC perform well due to large reservoir and small time step. The hybrid displays no benefit over the standalone RC or NGRC when using large amounts of training data, consistent with the results of Section \ref{sec:hybrid_improvement}. However, at lower to intermediate training data amounts the data efficiency of the NGRC is evident, performing much better than the RC and hybrid for $n_{\text{train}} \sim 10^{2.5}$ to $\sim 10^3$. 

In summary, the hybrid RC-NGRC approach can offer substantial improvements in predictive performance even when training on a limited amount of data, but is not beneficial if a system is well-suited to a standalone NGRC. 

\subsection{Hybrid RC-NGRC performance on other chaotic systems}
\label{sec:hybrid_other}

\begin{figure}
    \includegraphics[width=0.5\textwidth]{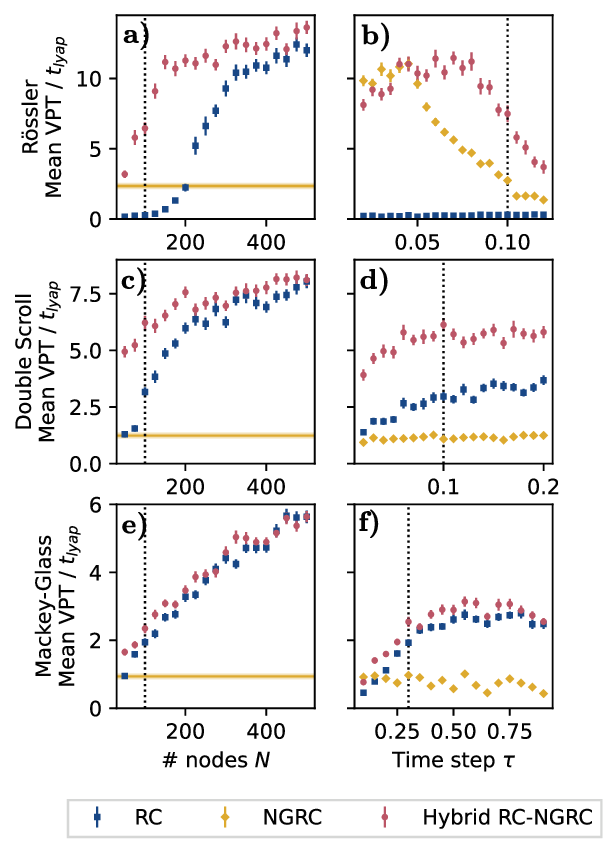}
    \caption{\label{fig:other_systems} Mean valid prediction times versus number of reservoir nodes $N$ and versus time step $\tau$ for several chaotic systems. Dotted vertical lines indicate the reservoir size/time step used in the adjacent plot in the same row; all other hyperparameters are as in Table \ref{tab:hybrid_hyperparameters}. 
    \textbf{a-b)} For the Rössler system, for which the essential nonlinearities are contained in the NGRC feature vector, the hybrid RC-NGRC has the greatest advantage when the RC is weak due to small size and the NGRC is weak due to large time step, just as in the Lorenz system. 
    \textbf{c-d)} For the Double Scroll system \cite{chang.etal_1998}, our NGRC implementation is weak due to the feature vector not containing the appropriate nonlinearities. Hybridizing with a weak RC shows substantial improvement, over a wide range of time steps. 
    \textbf{e-f)} For the Mackey-Glass system, the NGRC lacks sufficient utility, giving the hybrid only marginal improvement over the RC alone.
    }
\end{figure}

We now test the hybrid RC-NGRC approach for forecasting a few other prototypical chaotic systems \cite{gilpin_2023a}. Following our investigation in the Lorenz system, we test how the mean valid prediction times (VPTs) of RC, NGRC, and hybrid RC-NGRC forecasts vary with reservoir size $N$ and time step size $\tau$, while keeping the other hyperparameters fixed as in Table \ref{tab:hybrid_hyperparameters}. We further demonstrate that hybridizing a weakly performing RC (due to small reservoir size) with a weakly performing NGRC (either due to large time step, as in our forecasts of the Rössler and Lorenz systems; or due to inaccurate nonlinearities in the NGRC feature vector, as in our forecasts of the Double Scroll system \cite{chang.etal_1998}) yields substantial benefit in terms of short term prediction quality and long term climate replication over the RCs and NGRCs alone. However, we observe that if the NGRC is too weak, as we see in the Mackey Glass system, the hybrid approach offers just a modest benefit over a reservoir alone.

The chaotic Rössler system \cite{rossler_1976} has only one quadratic nonlinear term in its governing differential equations, so just like for the Lorenz system our quadratic NGRC feature vector contains all appropriate nonlinearities of the system. Plots of mean VPT vs. $N$ and $\tau$ (Figure \ref{fig:other_systems}a-b)) echo those of the Lorenz system (Figure \ref{fig:VPT_vs_size_and_timestep}), showing that the hybrid RC-NGRCs have the greatest relative advantage for short term prediction when RCs are weak due to small size and NGRCs are weak due to large time step. Under these conditions, our hybrid RC-NGRC also achieves improved climate replication, as shown by a much lower mean normalized map error than the RC or NGRC (Table \ref{tab:map_errs}).

The chaotic double-scroll electronic circuit introduced in Ref. \cite{chang.etal_1998} has governing differential equations that contain hyperbolic sine terms that are not represented in our quadratic NGRC feature vector. Although it is possible to craft well-performing NGRCs for this system \cite{gauthier.etal_2021} by adding terms to the feature vector, here we stick with our naive quadratic implementation, yielding a poorly performing NGRC. Figure \ref{fig:other_systems}c-d) shows that hybrids of these weak NGRCs with small RCs can provide substantial benefits over either of the two components. This benefit is found over a wide range of time step sizes. Again, we see strong climate replication in our hybrid RC-NGRC, evidenced by the mean normalized map error of the hybrid RC-NGRC being lower  than that of the RC and NGRC, as shown in Table \ref{tab:map_errs}.

However, hybridizing a small RC with an NGRC does not always provide substantial improvements in performance, as we observe in the Mackey-Glass system, for which our NGRC implementation is especially weak. The Mackey-Glass system \cite{mackey.glass_1977} presents a distinct challenge from the previous systems, as the governing differential equations include a time-delay term: the flow of the system at time $t$ depends not just on $\vb{u}(t)$ but also $\vb{u}(t-T)$, where here $T=2$. Even when adjusting the NGRC lookback time to approximately match the delay term ($s = 6$), we find our NGRC with quadratic nonlinearities struggles to make short-term accurate forecasts. Unlike for the previous systems, Figure \ref{fig:other_systems}e-f) shows that hybrid RC-NGRCs show only marginal improvement over RCs in terms of short term forecasting. (Note that we do not report a mean normalized map error for forecasts of the Mackey Glass system because doing so would require the sampling time step to match the integration time step. Due to the time delay term, integration of the governing differential equation using a time step $\tau_{\text{int}}$ requires a history of past states with time step $\tau_{\text{int}}$. Such a history is not provided in the prediction $\vb{v}$, which has time step $\tau \gg \tau_{\text{int}}$, making the calculation of Equation \ref{eqn:map_err_defn} impossible.)

We emphasize that blind application of the hybrid RC-NGRC approach to new dynamical systems may not in general give better performance than RC or NGRC alone. We expect the hybrid approach may be particularly beneficial when the NGRC is limited, e.g. by a large sampling time step or inaccurate nonlinearities, but still somewhat useful, and when one desires the computational efficiency of a small RC which by itself does not offer strong performance.

\section{Discussion and Conclusion}
\label{sec:conclusion}

We have introduced a hybrid RC-NGRC method for time series forecasting of dynamical systems. In the model chaotic systems we tested, we demonstrate that the hybrid method can make predictions that are both accurate in the short term and capture the system climate in the long term, even when the RC and NGRC components alone cannot. In addition, compared with its RC component, the hybrid is relatively insensitive to careful tuning of reservoir hyperparameters. We find that the hybrid RC-NGRC method holds the greatest advantage over its components when they both have only modest utility alone. In particular, we have shown that the hybrid offers strong performance when the RC is limited by small reservoir size and/or sub-optimal hyper parameters and the NGRC is limited because the training data is sparsely sampled (as in the Lorenz and Rossler systems) or the NGRC feature vector lacks the essential nonlinearities (as in the Double Scroll system). In these cases, we find that hybridizing a small RC with a weak NGRC provides the performance of a much larger RC, while offering substantial computational advantages. However, our results suggest that if the NGRC is too limited (as in the Mackey Glass system), our hybrid approach offers only modest benefits compared with a standalone RC. 

The performance improvements of the hybrid RC-NGRC arise from the complementary utility of its components. By contrast, for example, hybridizing two small RCs with different random network realizations does not provide the same benefit as seen here. We note that our hybrid approach has similarities to previous work on hybridizing an RC with a another machine learning model - Sparse Identification of Nonlinear Dynamical Systems (SINDy) \cite{koster.etal_2023}. However, a prominent difference in our approach that provides additional flexibility is that the terms of the NGRC feature vector themselves, rather than just the output of a model, are directly included in the hybrid feature vector. While this method of concatenating components is particularly natural in the case of RC-NGRC hybridization because both use ridge regression to fit the output weights, it can be adapted for hybridization of an RC with another ML-based model. For example, forming a hybrid of a RC and only those candidate terms that SINDy has identified as having non-zero coefficients could be an interesting avenue for further study.

In summary, we believe that the hybrid RC-NGRC scheme is an important step toward lightweight and flexible reservoir computing which leverages the computational efficiency of NGRCs while still maintaining the robustness of traditional RCs.

\section{Supplementary Material}
\label{sec:supplementary}

Supplementary materials can be found in a separate file.

\section{Acknowledgements}
\label{sec:acknowledgements}

We would like to thank Andrew Pomerance, Declan Norton, and Brian Hunt for their useful discussions about this manuscript. DA's contribution to this research was supported by NSF award PHY-2150399. MG and RC's contributions were supported by ONR award number N000142212656. 

\section{Author Declarations}
\label{sec:author_declarations}

All authors have no conflicts to disclose.

\textbf{Ravi Chepuri:} writing - original draft (lead); implementation/methodology (lead); visualization (equal), data analysis (equal) \textbf{Dael Amzalag:} data analysis (equal); visualization (equal); implementation/methodology (supporting) \textbf{Thomas Antonsen:} conceptualization and design (supporting); supervision (supporting) \textbf{Michelle Girvan:} conceptualization and design (lead); supervision (lead); editing (lead).

\section{Data Availability Statement}
\label{sec:data_availability}

The data in this study can be generated by running the publicly available code (see the code availability statement).

\section{Code Availability Statement}
\label{sec:code_availability}

All code is available under an MIT License on Github (\url{https://github.com/ravi-chepuri/hybrid_RC_NGRC}).

\bibliography{references.bib}

\end{document}


\title[Working title]{Supplementary Material: Hybridizing Traditional and Next-Generation Reservoir Computing to Accurately and Efficiently Forecast Dynamical Systems}

\author{R. Chepuri}
    
\author{D. Amzalag}

\author{T.M. Antonsen}

\author{M. Girvan}
    \email{girvan@umd.edu}

\date{\today}

\maketitle
In this supplemental material to our main text on ``Hybridizing Traditional and Next-Generation Reservoir Computing," we discuss noise regularization (Section \ref{sec:input_noise}), the role of memory (Section \ref{sec:memory}), and sensitivity to other hyperparameters (Section \ref{sec:hyperparams}).

\section{Noise Regularization}
\label{sec:input_noise}

In the main text, we add small-amplitude noise to the training data when feeding it into the RC, NGRC, and hybrid RC-NGRC (we still use noiseless data as training targets when fitting the output matrix). This input noise technique has been shown to promote climate stability of forecasts during autonomous prediction by mapping small perturbations off the attractor back onto it [1]. Here, we demonstrate that the central results of the paper still hold when using noiseless input data, though the input noise technique does confer some useful benefits for short term forecasting and climate replication.

\begin{figure}[h]
    \includegraphics[width=\textwidth]{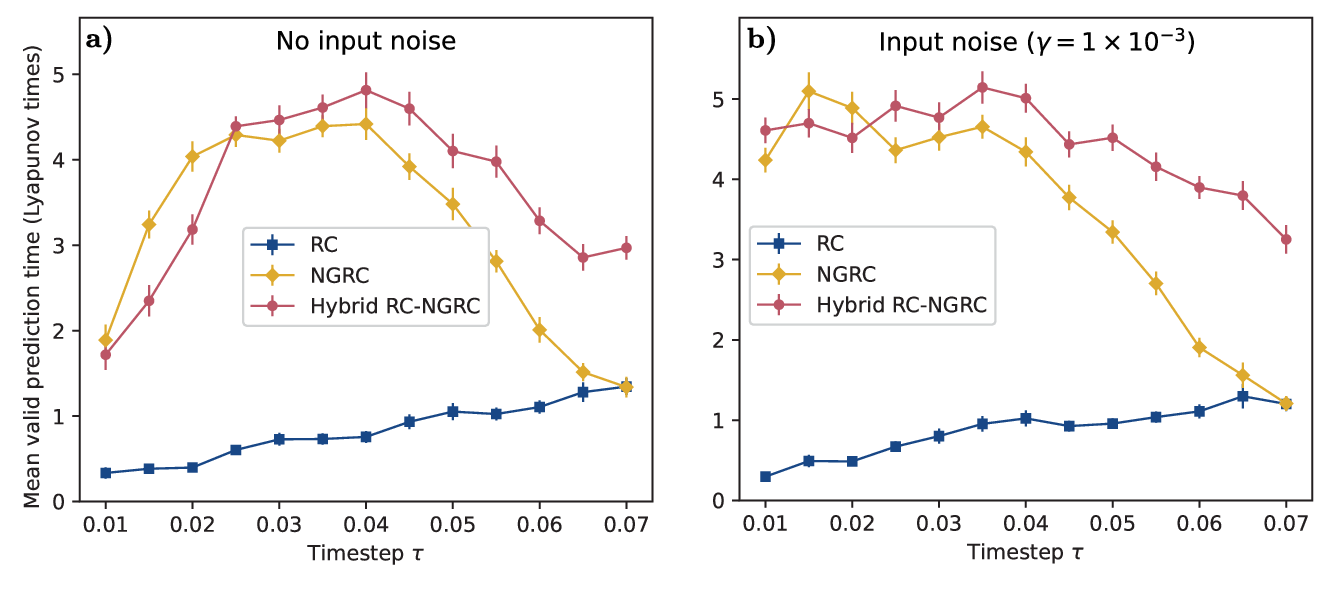}
    \caption{\label{fig:VPT_vs_timestep_nonoise} Mean valid prediction times for the Lorenz system versus time step size $\tau$ in the training data, with \textbf{a)} no input noise used and \textbf{b)} input noise standard deviation $\gamma = 1 \times 10^{-3}$ used. Regardless of noise, the hybrid RC-NGRC shows the greatest advantage in predictive power over the RC or NGRC alone when using a large time step. Reservoir size $N = 100$. Error bars: standard error of the mean (64 trials). Figure b) is repeated from Figure 4b) of the main text.}
\end{figure}

All results in the main text showing the advantage of the hybrid RC-NGRC over the RC and NGRC alone in terms of short term predictive power still hold with no input noise (noise standard deviation $\gamma = 0$). For example, the plot of VPT for predicting the Lorenz system vs. training data time step in Figure \ref{fig:VPT_vs_timestep_nonoise}a) shows the same qualitative behavior at large time steps as Figure 4b) of the main text (reshown in Figure \ref{fig:VPT_vs_timestep_nonoise}b) for ease of comparison). The key result that the hybrid RC-NGRC has better short term predictive performance of the Lorenz system compared witht an RC or NGRC alone, specifically at large time steps, is unchanged.

\begin{figure}[h]
    \includegraphics[width=\textwidth]{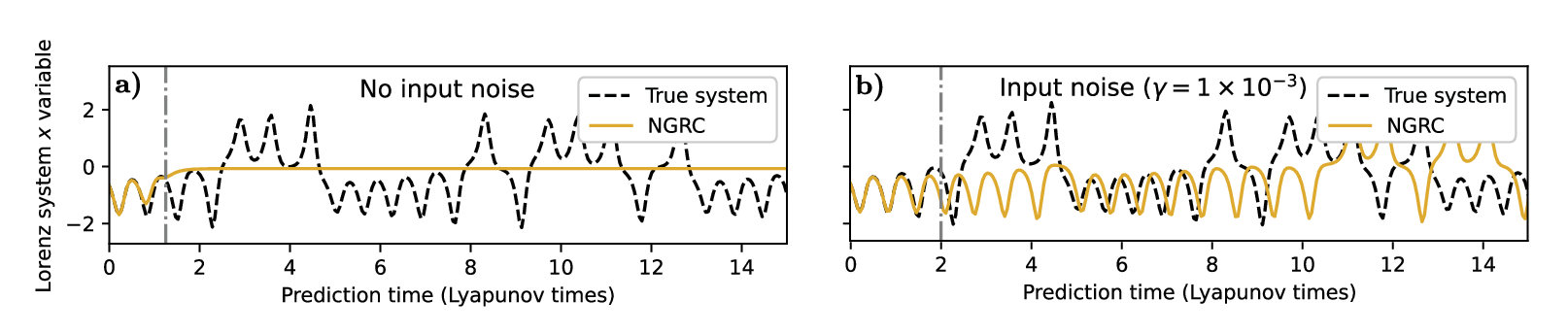}
    \caption{\label{fig:example_NGRC_prediction} Representative examples of autonomous NGRC predictions of the Lorenz system with large time step $\tau = 0.06$ with \textbf{a)} no input noise used and \textbf{b)} input noise standard deviation $\gamma = 1 \times 10^{-3}$ used (shown: $x$ component). }
\end{figure}

However, without the input noise technique, the climate replication abilities of the autonomous predictions are much worse, especially those of the NGRC. Figure \ref{fig:example_NGRC_prediction}a) shows a representative example of an NGRC autonomous prediction with no noise. The NGRC prediction quickly limits to a fixed point, failing completely to capture the climate; similar failures are observed in most trials across many initial conditions. The failure to capture climate is much more severe than when using the input noise technique as shown in Figure \ref{fig:example_NGRC_prediction}b), where the NGRC predictions track the true attractor for a longer time. If it occurs quickly enough, the sudden convergence of NGRC predictions to a fixed point can harm short term predictive power, for example contributing to a decreased mean VPT at small time steps in Figure \ref{fig:VPT_vs_timestep_nonoise}a). The hybrid RC-NGRC predictions also sometimes fail to capture the Lorenz attractor, in some trials limiting to a fixed point or limit cycle, in contrast with the main text where the hybrid RC-NGRC with input noise always captured the Lorenz system climate. Ref. [1] demonstrated that the input noise technique can stabilize autonomous predictions by preventing them from diverging numerically. Our findings suggest that in NGRCs, input noise is also helpful to prevent predictions from converging to a fixed point.

\begin{figure}[h]
    \includegraphics[width=\textwidth]{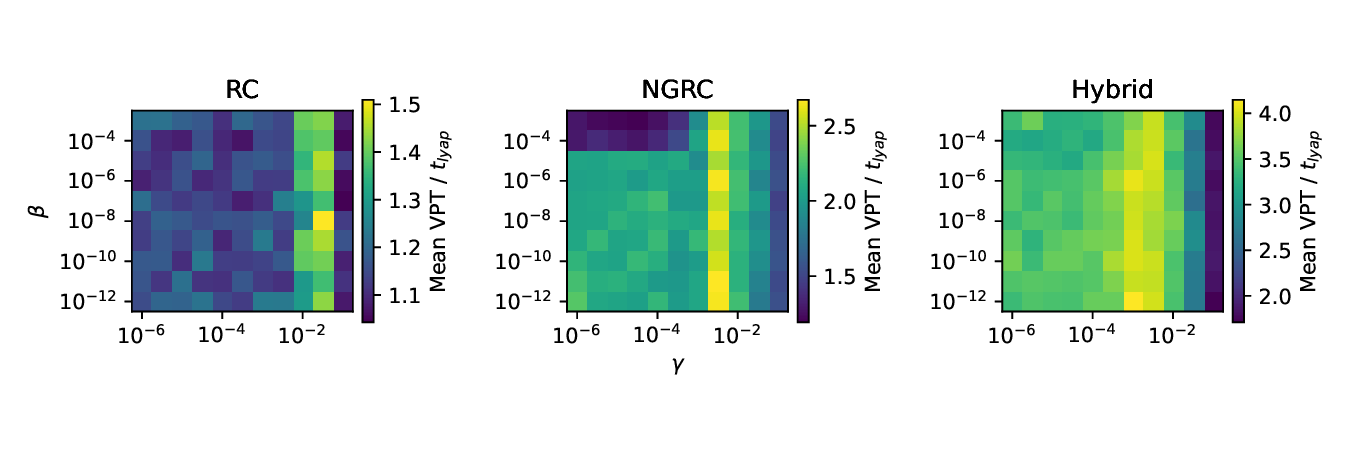}
    \caption{\label{fig:VPT_heatmaps} Mean valid prediction times for the Lorenz system versus input noise standard deviation $\gamma$ and Tikhnonov regularization parameter $\beta$. RC and NGRC have an optimal noise level for short term prediction, but the trend is not as clear for the hybrid RC-NGRC. (256 trials)}
\end{figure}

We also show that the benefits of input noise cannot be achieved by tuning the strength of ridge regression parameter in Figure \ref{fig:VPT_heatmaps}. The heat maps of Figure \ref{fig:VPT_heatmaps} show that both the RC and NGRC have optimal nonzero noise strengths for short term predictive performance. Figure \ref{fig:VPT_heatmaps} confirms this trend occurs over a wide range of the ridge regularization hyperparameter values $\beta$, showing that the benefits of noise training cannot be achieved just by tuning $\beta$. The hybrid RC-NGRC also appears to have an optimum noise strength, but trend is not as sharp as in RC and NGRC.

Note that the input noise technique used here is distinct from simply adding observational noise to the training data, as the train targets are still taken to be non-noisy values. The input noise technique is also distinct from using dynamical noise in the underlying system.

\section{The Role of Memory}
\label{sec:memory}

\begin{figure}[h]
    \includegraphics[width=0.5\textwidth]{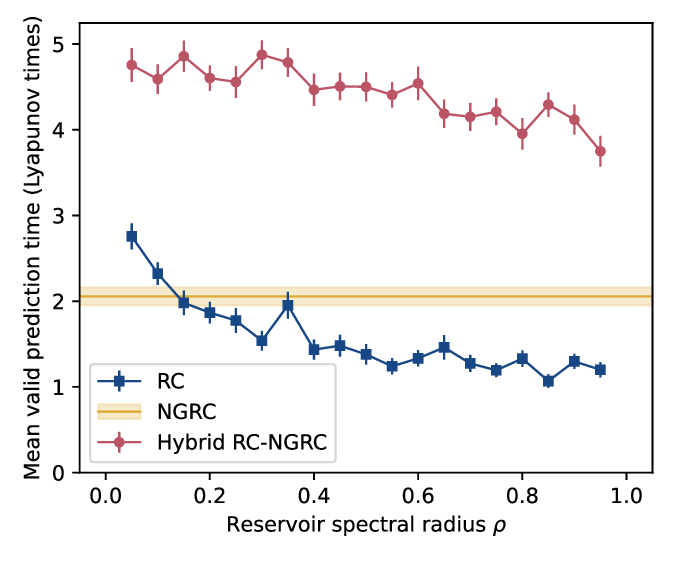}
    \caption{\label{fig:SI1} Mean valid prediction times for the Lorenz system vs. spectral radius of the reservoir. Other hyperparameters are listed in Table I of the main text. Error bars and band: standard errors of the mean (64 trials). }
\end{figure}

Typical RC implementations use a spectral radius $\rho$ slightly less than 1 to achieve the so-called ``echo-state property,'' endowing the reservoir with a memory of past states [2]. One might expect that the efficacy of the hybrid RC-NGRC is due to combining a long-memory RC with a short memory NGRC, allowing these two components to focus in a complimentary fashion on different time scale dynamics. However, here we show that this is not the case. For some tasks, such as the full-state Lorenz forecasting task considered in the main text, memory may not be useful or necessary--some results have shown that reservoirs with $\rho \approx 0$ can perform well for this task [3, 4]. In Figure \ref{fig:SI1}, we tune the spectral radius down to $\rho = 0.05$, yielding a memory time scale on the order of a single time step, and indeed we see strong RC predictive performance. Nonetheless, we see that the hybrid RC-NGRC maintains its advantage over the RC and NGRC at all $\rho$. In particular, the advantage of the hybrid RC-NGRC is not dependent on the two components having different memory time scales.

\begin{figure}[h]
    \includegraphics[width=0.5\textwidth]{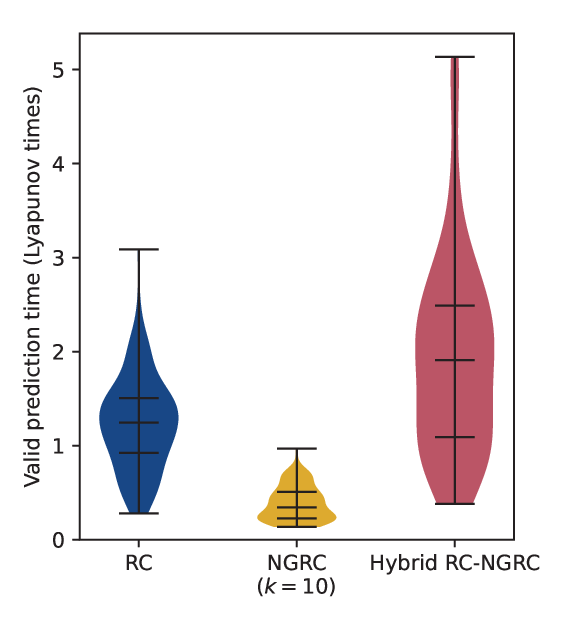}
    \caption{\label{fig:SI2} Valid prediction time distributions for the Lorenz partial state forecasting task (100 trials). The hybrid RC-NGRC offers a similar improvement for the partial state observation and forecasting as we saw for full-state observation and forecasting in the main text.}
\end{figure}

For other forecasting problems in which memory is important, such as partial-state observation and forecasting of the Lorenz system (given only a time series of $x$-components, forecast the future $x$-components), we find that the hybrid RC-NGRC provides a similar advantage to what we saw for full-state observation in the main text. An NGRC with even $k=10$ observational terms struggles with this task since the appropriate nonlinearities cannot be readily constructed from the data. An RC with $\rho=0.9$ performs slightly better. However, hybridizing with the poorly performing NGRC does appear to provide some advantage in terms of mean valid prediction times as shown in Figure \ref{fig:SI2} (hybrid: $1.91 t_{\text{lyap}}$, RC: $1.25 t_{\text{lyap}}$, NGRC: $0.34 t_{\text{lyap}}$).

\section{Sensitivity to Other Hyperparameters}
\label{sec:hyperparams}

\begin{figure}
    \centering
    \subfloat{\includegraphics[width=.5\textwidth]{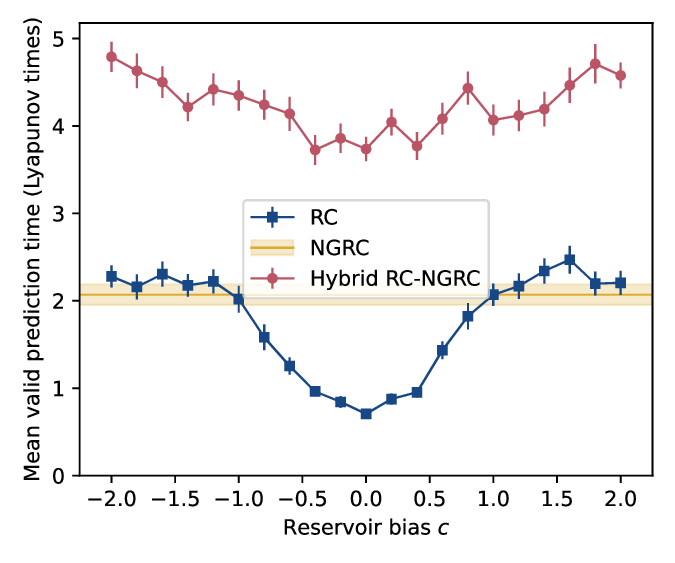} }
    \subfloat{\includegraphics[width=.5\textwidth]{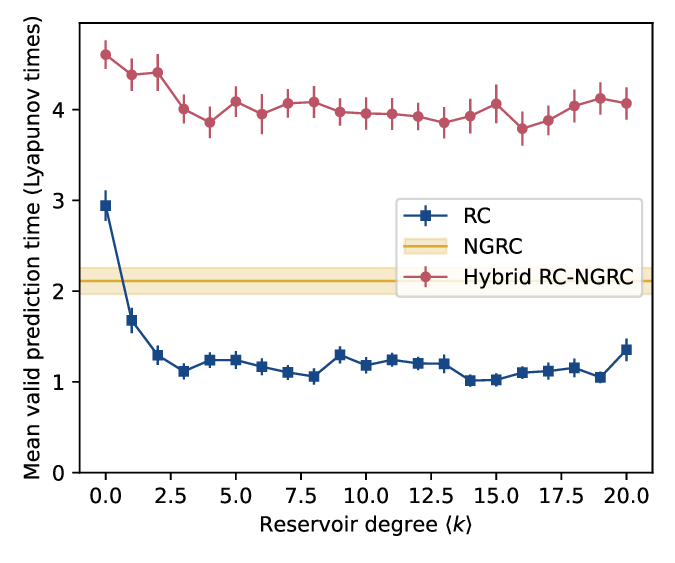} }
    \caption{\label{fig:VPT_vs_hyperparams} Mean valid prediction times for the Lorenz system vs. reservoir bias $c$, average degree $\langle k \rangle$, and regularization hyperparameter $\beta$. Other hyperparameters are listed in Table I of the main text. Error bars and band: standard errors of the mean (64 trials).}
\end{figure}

In Figure 5 of the main text we show that the hybrid RC-NGRC is much less sensitive to the input matrix scaling $\sigma$ of the reservoir than the RC alone. In Figure \ref{fig:VPT_vs_hyperparams} we show that similar trends as we vary two other hyperparameters: bias $c$ and the reservoir degree $\langle k \rangle$. We additionally note that the RC, NGRC, and hybrid RC-NGRC appear insensitive to the regularization hyperparameter $\beta$ (as illustrated in Figure \ref{fig:VPT_heatmaps}).

\vspace{5mm}
\hrule
\vspace{5mm}
[1] A. Wikner, J. Harvey, M. Girvan, B. R. Hunt, A. Pomerance, T. Antonsen, and E. Ott, Neural Networks \textbf{170}, 94 (2024).

[2] H. Jaeger, in \textit{GMD - Ger. Natl. Res. Inst. Comput. Sci.}, Vol. 148 (2001).

[3] A. Griffith, A. Pomerance, and D. J. Gauthier, Chaos: An Interdisciplinary Journal of Nonlinear Science \textbf{29}, 123108 (2019).

[4] L. Jaurigue, “Chaotic attractor reconstruction using small reservoirs – the influence of topology,” (2024), arxiv:2402.16888 [math-ph, physics:nlin, physics:physics].
